\documentclass[11pt]{article}

\usepackage[final]{acl}

\usepackage{times}
\usepackage{latexsym}

\usepackage{amsmath}
\usepackage{multirow}
\usepackage{booktabs}
\usepackage{cleveref}
\usepackage{xcolor}
\usepackage{soul}

\usepackage[T1]{fontenc}

\usepackage[utf8]{inputenc}

\usepackage{microtype}

\usepackage{inconsolata}

\usepackage{graphicx}

\newcommand{\hlgreen}[1]{{\sethlcolor{green!20}\hl{#1}}}
\newcommand{\hlred}[1]{{\sethlcolor{red!20}\hl{#1}}}
\newcommand{\hlorange}[1]{{\sethlcolor{orange!20}\hl{#1}}}

\title{Leveraging Low-Level Symbolic Competences \\ for Unsupervised Grounding in Hallucination Detection}

\author{Renato Vukovic, Hsien-chin Lin, Carel van Niekerk, Benjamin Ruppik, \\{\bf Michael Heck, Shutong Feng, Nurul Lubis, Milica Ga\v{s}i\'{c}} \\
Heinrich Heine University Düsseldorf, Germany\\
\texttt{\small{\{revuk100,linh,ruppik,heckmi,fengs,lubis,gasic\}@hhu.de}}\\
\texttt{\small{academic@carelvniekerk.com}}
}

\begin{document}
\maketitle
\begin{abstract}
Hallucination—where a language model generates outputs that are factually incorrect or unsupported by the source—is a major challenge for both prompted and fine-tuned language models. Detecting hallucinations is difficult due to the opaque reasoning processes of LLMs, which often provide little insight into why a model's output may be inaccurate. 

In this work, we investigate whether an LLM can use an alternative, low level, symbolic competence such as SQL for unsupervised hallucination detection in some high level task.
For this, we make an LLM build an SQL database from reference documents. 
This SQL database is then used for reasoning over the reference and the sampled response in a hallucination detection pipeline that is grounded in the database, thereby providing a neurosymbolic checkup.

On RAGTruth and DiaHalu hallucination detection datasets, we find that our approach improves on direct prediction and competes with state-of-the-art hallucination detection methods, while not requiring domain-specific fine-tuning.
Instead it relies on a low-level general competence already present in LLMs.
This warrants further investigation of low-level LLM competences in neurosymbolic approaches.
\footnote{Code is available under \url{https://github.com/renatovukovic/teqhallu}}
\end{abstract}

\section{Introduction}

Large language models (LLMs) excel across natural language processing tasks due to large-scale pretraining and instruction tuning \cite{brown2020language,ouyang2022training}.
Consequently, they form the backbone of modern AI systems by enabling fluent interactions.

Still, LLMs frequently generate factually incorrect outputs unsupported by the source context, a phenomenon known as \emph{hallucination} \cite{ji2023hallucinationsurvey,huang2024hallucinationsurvey}.
These hallucinations pose significant risks in high-stakes domains like medicine and law, where misinformation undermines decision-making \cite{weidinger2022riskLMs}.

\begin{figure}[t]
    \centering
    \includegraphics[width=0.95\linewidth]{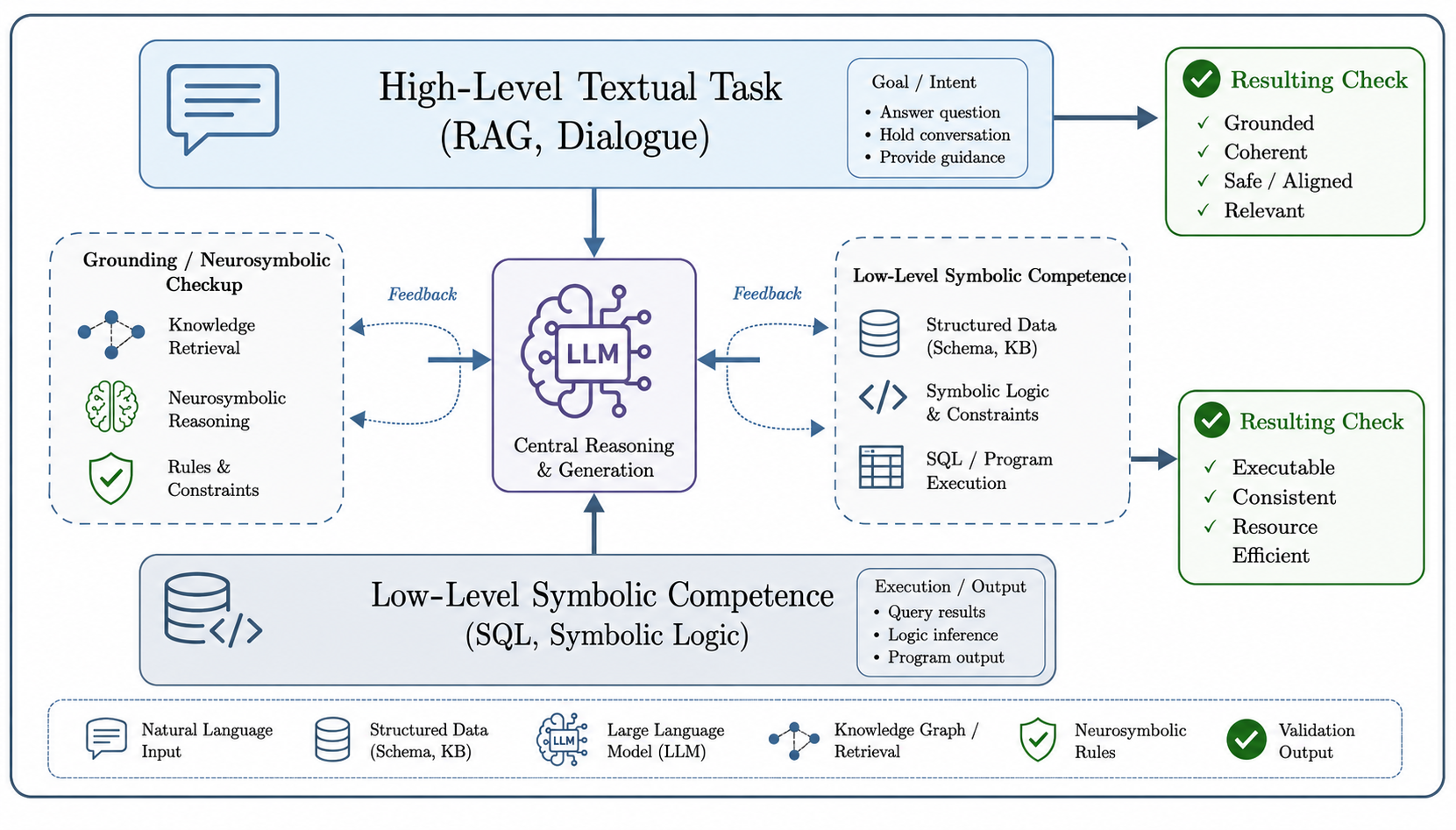}
    \caption{Grounding hallucination detection using low-level symbolic competences.}
    \label{fig:intro_figure}
    \vspace{-15pt}
\end{figure}

Detecting hallucinations remains challenging because the opaque reasoning processes of LLMs provide little insight into why an output is inaccurate \cite{doshi2017towards}.
While existing detection methods achieve strong performance, they often act as black boxes \cite{manakul-etal-2023-selfcheckgpt,farquhar2024detecting,chen2024inside}.
This opacity persists even in modern retrieval-augmented generation \citep[RAG;][]{lewis2020retrieval} systems, offering limited interpretability regarding unsupported claims \cite{niu-etal-2024-ragtruth,song-etal-2024-rag,kovacs-recski-2025-lettucedetect,dubanowska-etal-2025-representation}.

\begin{figure*}[t]
    \centering
    \includegraphics[width=\linewidth]{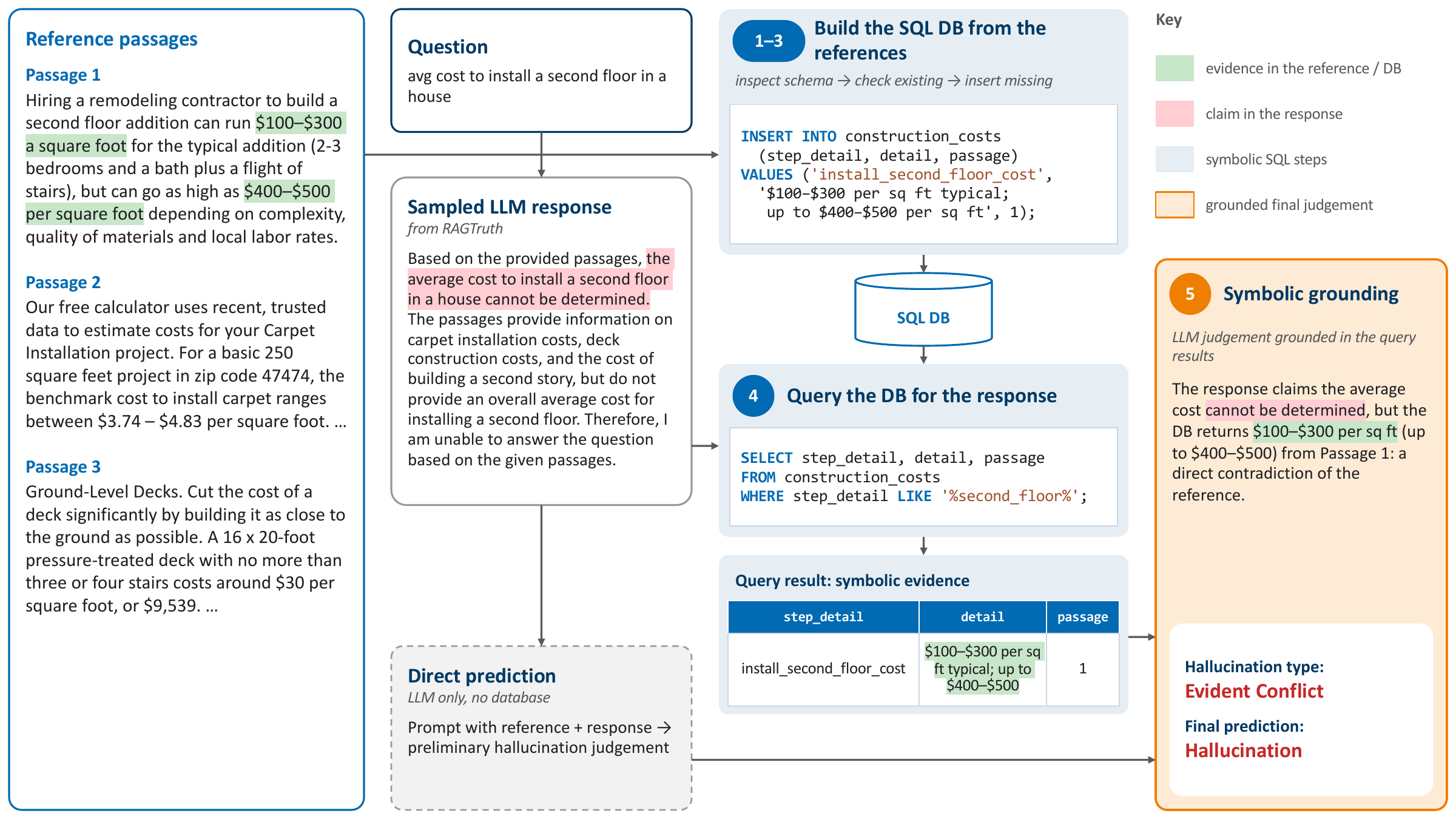}
    \caption{Text-to-SQL grounded hallucination detection with an example from the RAGTruth QA task.}
    \label{fig:main}
    \vspace{-12pt}
\end{figure*}

To improve robustness, recent work explores grounding LLM reasoning in structured or symbolic representations \cite{west-etal-2022-symbolic,yang2025surveryknowledgedistillation}.
Furthermore, \citet{dennett1987intentional} argues that explicit task comprehension is not strictly necessary for task competence (\Cref{fig:intro_figure}).
Building on this, the TeQoDO pipeline \citep{vukovic2025text} demonstrates that an LLM's lower-level competence in SQL can be leveraged to construct a reliable knowledge base for task-oriented dialogue \citep{eric-etal-2020-multiwoz}.

In this work, we investigate whether an LLM can utilise this lower-level competence for unsupervised hallucination detection.
We propose \textbf{TeQHallu}, a \textbf{T}ext-to-S\textbf{Q}L pipeline for \textbf{Hallu}cination detection that builds on TeQoDO.
Our approach prompts an LLM to build a relational SQL database from reference documents, which is then used to reason over both the reference and the response.
Grounding the detection pipeline in executable SQL queries provides a neurosymbolic \cite{garcez2019neurosymbolic,garcez2023neurosymbolic} checkup without requiring domain-specific fine-tuning.

Our contributions can be summarised as follows:
\begin{itemize}
    \item We introduce TeQHallu, an unsupervised text-to-SQL method that leverages low-level LLM competence to structure reference knowledge for hallucination detection.
    \item We demonstrate that grounding detection in an SQL database provides transparent, inspectable reasoning traces compared to black-box approaches.
    \item We evaluate our pipeline on the RAGTruth and DiaHalu benchmarks, performing competitively against state-of-the-art methods without domain-specific fine-tuning.
    \item We highlight neurosymbolic checkups via lower-level, task-agnostic competences as a promising avenue for improving LLM reliability.
\end{itemize}

\section{Related Work}

\subsection{Hallucination Detection}

Hallucination detection identifies factually incorrect or unsupported model outputs to enhance LLM reliability \cite{ji2023hallucinationsurvey, huang2024hallucinationsurvey}.
Early strategies primarily relied on sequence-level metrics and heuristic rules to flag improbable generations \cite{lin-etal-2022-truthfulqa}.

Recent methods evaluate internal model states and self-consistency, using sampling-based consistency \cite{manakul-etal-2023-selfcheckgpt}, hidden layer signals \cite{li2025llmhallucination}, or semantic entropy \cite{farquhar2024detecting} to estimate uncertainty.
In retrieval-augmented generation (RAG) contexts, cross-referencing model outputs with external knowledge serves as the primary safeguard \cite{lewis2020retrieval, niu-etal-2024-ragtruth}.
However, these techniques typically function as ``black boxes'', yielding binary labels without interpretable reasoning or granular justifications \cite{sriramanan2024llmcheck}.




\subsection{Text-to-SQL and Structured Reasoning}

Text-to-SQL methodologies allow LLMs to use executable queries as a structured reasoning tool, enhancing factual reliability \cite{gao2024text2sqlllm,liu2025text2sqlsurvey}.
Advanced frameworks like DIN-SQL decompose complex queries into verifiable sub-tasks to improve generation accuracy \cite{pourreza2023din}.
Dialogue ontology construction methods tend to focus on individual stages of the process, such as extracting ontology relations between existing concepts \citep{vukovic-etal-2024-dialogue}.
\citet{vukovic2025text} demonstrate that text-to-SQL can be used to construct task-oriented dialogue ontologies from scratch.
However, these approaches lack a downstream application for the generated knowledge base.
Our framework addresses this gap by repurposing these structured ontologies specifically for hallucination detection.

Neurosymbolic methods integrate language models with symbolic constraints to improve prediction quality, robustness, and interpretability \cite{princis2025enhancing}.
Grounding reasoning in executable, symbolic formats yields both tangible performance gains and superior explainability.

However, existing text-to-SQL systems primarily focus on query accuracy rather than hallucination verification \cite{pourreza2023din,gao2024text2sqlllm}.
The deployment of text-to-SQL pipelines as explicit evaluation mechanisms in dialogue systems remains largely understudied.
This motivates the TeQHallu approach, which leverages SQL grounding to provide a factual anchor and a transparent audit trail for model responses.

\section{The TeQHallu Framework}

TeQHallu is a neurosymbolic framework for unsupervised hallucination detection in retrieval-augmented generation settings.
It combines probabilistic language generation with deterministic symbolic verification.
Rather than relying solely on neural reasoning, it converts retrieved text into a structured, queryable relational representation.
TeQHallu is illustrated in \Cref{fig:main} and the full prompt in Appendix \ref{sec:appendix:prompt}.

\subsection{Incremental Database Construction via Multi-step Prompting}

TeQHallu constructs a relational database from retrieved passages through multi-step prompting.
The process begins by prompting the LLM to inspect the current schema using metadata queries such as \texttt{PRAGMA table\_info}.
This provides symbolic awareness of existing tables, columns, and stored information.
The model then analyses retrieved passages to identify information absent from the database.
Instead of rebuilding the schema, the framework extracts the symbolic delta between textual evidence and the current representation.

Based on this analysis, the LLM generates database operations to insert new information or extend the schema.
The prompting encourages the conversion of natural language variables, measurements, entities, and qualifiers into relational tuples.
In our example, the model extracts second-floor cost information, storing numerical ranges with semantic labels and source provenance.
The statement that second-floor construction typically costs \$100--\$300 per square foot, rising to \$400--\$500, is encoded as structured symbolic evidence.
This incremental strategy maintains a high-fidelity relational abstraction of retrieved context without domain-specific supervision.

\subsection{Claim Grounding via Symbolic Retrieval}

After database construction, the framework grounds claims through symbolic retrieval.
The generated response is first analysed to isolate claims requiring factual verification.
The system then prompts the LLM to generate targeted \texttt{SELECT} queries conditioned on the original question and response.
These prompts guide retrieval of database entries most relevant to validation.
The resulting SQL queries are executed to obtain symbolic evidence derived directly from the source passages.
This transforms verification from implicit reasoning into an explicit retrieval task grounded in database operations.

In the example, the response claims that the average cost of installing a second floor cannot be determined from the passages.
The SQL retrieval step instead returns stored cost information from the reference material.
The retrieved evidence includes a typical range of \$100--\$300 per square foot and estimates up to \$400--\$500.
The database therefore contains sufficient information to answer the query.
Because the response asserts informational absence despite available evidence, the framework detects a contradiction between the response and symbolic retrieval outputs.
The final assessment is thus grounded in the database rather than solely in the model's internal parameters.

\subsection{Neurosymbolic Checkup and Alignment}

The final stage performs a neurosymbolic checkup that reconciles neural predictions with symbolic retrieval results.
An initial hallucination prediction is generated through prompting based only on the LLM's internal reasoning.
This serves as a neural-only baseline judgement.
A second prompting stage then provides SQL retrieval outputs, interpreted database summaries, and relevant textual context.
The model is instructed to compare its preliminary judgement with the retrieved evidence and justify its final decision.

The framework evaluates whether a response is fully supported, partially supported, unsupported, or contradicted by database results.
In \Cref{fig:main}, the response states that the average installation cost cannot be determined from the passages.
This directly conflicts with structured evidence retrieved from the relational database.
The framework therefore classifies the response as an \textit{Evident Conflict}.
The final output includes both the hallucination prediction and a structured explanation grounded in SQL retrieval results.
This multi-stage procedure produces an inspectable reasoning trace with greater transparency than purely neural detection approaches.

\section{Experiments}

\subsection{RAGTruth Dataset}

We evaluate TeQHallu on the RAGTruth dataset \cite{niu-etal-2024-ragtruth}, which contains naturally occurring hallucinations across three distinct retrieval-augmented tasks: question answering \cite[MS MARCO;][]{Bajaj2016Msmarco}, data-to-text generation \cite[Yelp;][]{YelpOpenDataset}, and summarisation \cite[CNN/Daily Mail;][]{see-etal-2017-get}.
Each test set contains 900 document-response pairs.
The evaluation focus centers on model-generated responses sampled from a diverse array of foundational architectures, including GPT-3.5 \cite{openai_gpt35_2023}, GPT-4 \cite{openai_gpt4_2023}, and LLaMA variants \cite{touvron2023llama}.
The dataset captures errors regarding the retrieved document, categorised into four fine-grained types: evident conflict, subtle conflict, introduction of baseless information, and subtle introduction of baseless information.
Each response is manually annotated, and following the protocol in \cite{niu-etal-2024-ragtruth}, we evaluate binary response-level hallucination detection to determine whether a hallucination is present or not.
We report precision, recall, and F1-score for each individual task alongside macro-averaged results across the entire benchmark.

\subsection{DiaHalu Dataset: TOD Portion}

To assess performance on multi-turn dialogue data, we utilise the task-oriented dialogue (TOD) portion of the DiaHalu dataset \citep{chen-etal-2024-diahalu}.
This segment is anchored by structured database intents from MultiWOZ 2.1 \citep{eric-etal-2020-multiwoz} and DSTC 1.0 \citep{williams-etal-2013-dialog}, categorising hallucinations into non-factual, incoherence, irrelevance and over-reliance.
Conversations were generated using a self-chat framework with GPT-3.5 \citep{openai_gpt35_2023} and GPT-4 \citep{openai_gpt4_2023} based on distinct prompts for user goals and system responses.
The final expert-annotated corpus comprises 210 dialogues, consisting of 87 MultiWOZ-based dialogues, 56 DSTC 1.0-based dialogues, and 67 conversations based on synthetic GPT-4 user intents.

\begin{table*}[ht!]
\centering
\resizebox{1.\linewidth}{!}{
\small
\begin{tabular}{@{\hspace{0pt}}l@{\hspace{1pt}}|@{\hspace{2pt}}ccc@{\hspace{2pt}}|@{\hspace{2pt}}ccc@{\hspace{2pt}}|@{\hspace{2pt}}ccc@{\hspace{2pt}}|@{\hspace{2pt}}ccc@{\hspace{1pt}}}
\hline
\multirow{2}{*}{Methods} & \multicolumn{3}{c@{\hspace{2pt}}|@{\hspace{2pt}}}{Question Answering} & \multicolumn{3}{c@{\hspace{2pt}}|@{\hspace{2pt}}}{Data-to-text Writing} & \multicolumn{3}{c@{\hspace{2pt}}|@{\hspace{2pt}}}{Summarization} & \multicolumn{3}{c}{Overall Macro} \\
 & Precision & Recall & F1 & Precision & Recall & F1 & Precision & Recall & F1 & Precision & Recall & F1 \\
\hline
\textit{\cite{niu-etal-2024-ragtruth} initial baselines} & & & & & & & & & & & & \\
Prompt$_{\text{gpt-3.5-turbo}}$ & 18.8 & 84.4 & 30.8 & 65.1 & 95.5 & 77.4 & 23.4 & 89.2 & 37.1 & 35.8 & 89.7 & 48.4 \\
Prompt$_{\text{gpt-4-turbo}}$ & 33.2 & 90.6 & 45.6 & 64.3 & 100.0 & 78.3 & 31.5 & 97.6 & 47.6 & 43.0 & 96.1 & 57.2 \\
SelfCheckGPT$_{\text{gpt-3.5-turbo}}$ & 35.0 & 58.0 & 43.7 & 68.2 & 82.8 & 74.8 & 31.1 & 56.5 & 40.1 & 44.8 & 65.8 & 52.9 \\
LMvLM$_{\text{gpt-4-turbo}}$ & 18.7 & 76.9 & 30.1 & 68.0 & 76.7 & 72.1 & 23.3 & 81.9 & 36.2 & 36.7 & 78.5 & 46.1 \\
Finetuned Llama-2-13B & 61.6 & 76.3 & 68.2 & 85.4 & 91.0 & 88.1 & 64.0 & 54.9 & 59.1 & 70.3 & 74.1 & 71.8 \\
\midrule
\textit{SOTA fine-tuned baselines} & & & & & & & & & & & & \\
RAG-HAT (LLaMA-3-8B) \citep{song-etal-2024-rag} & 76.5 & 73.1 & 74.8 & 92.9 & 90.3 & 91.6 & 77.7 & 59.8 & 67.6 & 82.4 & 74.4 & 78.0 \\
lettuceedetect-base-v1 (150M)\citep{kovacs-recski-2025-lettucedetect} & 60.6 & 71.3 & 65.5 & 89.3 & 86.5 & 87.9 & 53.9 & 47.5 & 50.5 & 67.9 & 68.4 & 68.0 \\
lettuceedetect-large-v1 (396M) \citep{kovacs-recski-2025-lettucedetect} & 65.9 & 75.0 & 70.2 & 90.5 & 86.7 & 88.5 & 64.0 & 55.9 & 59.7 & 73.5 & 72.5 & 72.8 \\
ReDeEP$_\text{LLaMA-2 7B Chat}$\citep{SunEtAl2025_ReDeEP} & 45.3 & 46.1 & 45.7 & 79.3 & 74.8 & 77.0 & 48.4 & 29.4 & 36.6 & 57.7 & 50.1 & 53.1 \\
ReDeEP$_\text{LLaMA-2 13B Chat}$\citep{SunEtAl2025_ReDeEP} & 31.0 & 75.0 & 43.9 & 93.6 & 94.9 & 94.2 & 25.0 & 78.8 & 38.0 & 49.9 & 82.9 & 58.7 \\
ReDeEP$_\text{LLaMA-3 8B Instruct}$\citep{SunEtAl2025_ReDeEP} & 55.1 & 75.6 & 63.8 & 87.1 & 76.5 & 81.5 & 24.8 & 80.7 & 37.9 & 55.7 & 77.6 & 61.0 \\
\midrule
Direct Prediction Baseline$_\text{Gemini-2.5-Flash}$ & 30.1 & 97.5 & 45.9 & 76.6 & 95.5 & 85.0 & 38.0 & 93.1 & 54.0 & 48.2 & 95.4 & 61.6 \\
Multi-step Atomic Fact Checking \cite{min-etal-2023-factscore}$_\text{Gemini-2.5-Flash}$ & 35.0 & 92.5 & 50.8 & 79.0 & 94.5 & 86.1 & 45.9 & 88.2 & 60.4 & 53.3 & 91.7 & 65.7 \\
JSON Keyvalue Extraction$_\text{Gemini-2.5-Flash}$ & 33.3 & 95.6 & 49.4 & 81.1 & 95.5 & 87.7 & 41.9 & 93.1 & 57.8 & 52.1 & 94.8 & 64.9 \\
\midrule
\textbf{TeQHallu (ours)}$_\text{Gemini-2.5-Flash}$ & 42.8 & 92.5 & \textbf{58.5} & 86.8 & 90.0 & \textbf{88.4} & 55.5 & 84.3 & \textbf{66.9} & 61.7 & 88.9 & \textbf{71.3} \\
\hline
\end{tabular}
}
\caption{The response-level hallucination detection performance for each baseline method and different models, with overall macro averages over the different tasks from \citet{niu-etal-2024-ragtruth}. \citet{song-etal-2024-rag,kovacs-recski-2025-lettucedetect} are fine-tuned methods. ReDeEP results with the token-level method taken from \citet{dubanowska-etal-2025-representation}. \textbf{Bold} indicates the best F1-score for unsupervised methods.}
\label{tab:main_results}
\vspace{-10pt}
\end{table*}

\subsection{Implementation Details}

Our framework utilises \textit{Gemini 2.5 Flash} \cite{google2025gemini} via the Google Vertex AI API\footnote{\url{https://cloud.google.com/vertex-ai}} as the main model for both direct hallucination prediction and text-to-SQL grounded reasoning, with open-weight model configurations deferred to \Cref{experiments:subsec:open} for clarity of the main results.
Gemini 2.5 Flash is a general-purpose Mixture-of-Experts \cite[MoE;][]{Jacobs1991AdaptiveMixtures,Shazeer2017OutrageouslyLargeMoE} model optimised for computational efficiency and robust semantic reasoning.
The TeQHallu framework adopts the structured database construction, prompting strategy, and SQL interpretation logic established by the TeQoDO framework \cite{vukovic2025text} to maintain consistency in database synthesis.
For the DiaHalu dataset evaluation, the underlying databases are constructed directly from the text of the multi-turn dialogues under inspection.

\subsection{Baselines}

\paragraph{Initial RAGTruth Baselines}
\citet{niu-etal-2024-ragtruth} established the benchmark using both prompt-based and fine-tuned models.
Zero-shot configurations of GPT-3.5 and GPT-4 establish a baseline performance floor, while SelfCheckGPT \cite{manakul-etal-2023-selfcheckgpt} identifies factual inconsistencies via stochastic sampling across multiple model generations.
The LMvLM framework \cite{cohen-etal-2023-lm} applies a cross-examination strategy, utilising an external LLM to generate verification questions targeting specific response claims.
Furthermore, a fine-tuned Llama-2-13B model \cite{touvron2023llama} serves as a supervised baseline, illustrating the performance advantages of domain-specific training over general-purpose prompting.

\paragraph{SOTA Fine-tuned Baselines}
We compare TeQHallu against several state-of-the-art architectures optimized for robust hallucination detection.
RAG-HAT \cite{song-etal-2024-rag} utilizes Hallucination-Aware Tuning to generate binary labels and error descriptions for Direct Preference Optimisation \cite{rafailov2024direct}.
LettuceDetect \cite{kovacs-recski-2025-lettucedetect} employs a lightweight ModernBERT-based encoder \cite{warner-etal-2025-smarter} for efficient, span-level classification with a reduced parameter footprint.
ReDeEP \cite{SunEtAl2025_ReDeEP} decouples external context reliance from internal parametric knowledge by analyzing attention layers across various model scales.

\paragraph{Direct Prediction Baseline}
To isolate the impact of our symbolic reasoning layer, we evaluate a direct prediction baseline utilising Gemini 2.5 Flash.
This baseline employs zero-shot prompting to classify responses as faithful or hallucinated based on the reference text while providing a natural language explanation.
It provides the neural baseline, measuring performance based solely on internal reasoning.

\paragraph{Multi-Step and Structured Baselines.}
To isolate whether performance improvements stem specifically from relational SQL representations or simply from multi-step reasoning and information decomposition, we compare TeQHallu against two non-SQL verification paradigms. 
First, we implement an \textit{Atomic Claim-Level Decomposition} baseline following \cite{min-etal-2023-factscore} and \cite{dhuliawala-etal-2024-chain}, which breaks model responses into atomic natural-language claims and evaluates each claim step-by-step against the reference text without structured schemas. 
Second, we construct a \textit{Semi-Structured Key-Value} baseline that parses reference context into JSON key-value pairs (\texttt{\{"attribute": "value"\}}) and performs direct attribute lookups, allowing us to analyse whether relational SQL operations provide distinct advantages over simple structured representations.

\paragraph{DiaHalu Baselines}
We evaluate several state-of-the-art LLMs and specialised detection frameworks using results directly from the DiaHalu publication \citep{chen-etal-2024-diahalu}.
Open-source baselines include LLaMa-30B \citep{touvron2023llama} and Vicuna-33B \citep{chiang-etal-2023-vicuna}, which is fine-tuned for human-like dialogue.
Closed-source models comprise Gemini 1.5 Pro \citep{anil-etal-2023-gemini} and GPT-4 \citep{openai_gpt4_2023}, which provides a high-performance ceiling for complex logic.
Specialised detection methods include SelfCheckGPT \citep{manakul-etal-2023-selfcheckgpt} and FOCUS \citep{zhang-etal-2023-enhancing-uncertainty}, which detects hallucinations based on uncertainty.
We incorporate the official DiaHalu results across both one-shot and Chain-of-Thought \citep[CoT;][]{kohima2022cot_zeroshot} prompting configurations.

\subsection{Results}

\Cref{tab:main_results} presents response-level hallucination detection performance on the RAGTruth benchmark across question answering, data-to-text, and summarisation tasks.
Our proposed \textit{TeQHallu} method significantly outperforms all other unsupervised baselines, including SelfCheckGPT and the zero-shot \textit{Direct Prediction Baseline}, which consistently suffer from low precision and a tendency to over-predict hallucinations.
Crucially, by utilising structured verification to eliminate false positives, TeQHallu also outpaces the supervisedly fine-tuned Llama-2-13B model and performs almost on par with heavily supervised state-of-the-art (SOTA) systems like LettuceDetect-large-v1.

While the fine-tuned RAG-HAT model delivers the peak macro F1-score and remains the only configuration to outpace our framework, TeQHallu is highly promising as it achieves this competitive performance entirely out-of-the-box.
Ultimately, our approach proves that explicit symbolic reasoning can effectively compensate for the absence of parameter optimisation, delivering supervised-level accuracy without requiring any domain-specific fine-tuning.


Furthermore, evaluating multi-step non-SQL paradigms shows that both atomic claim-level decomposition and semi-structured key-value extraction improve upon direct prediction, confirming the benefit of multi-step information decomposition.
However, TeQHallu substantially outpaces both multi-step baselines across all three tasks, achieving a superior overall macro F1-score.
This advantage is driven by a notable increase in precision, demonstrating that relational SQL representations provide a far more robust mechanism for filtering false positives than flat text claims or simple key-value lookups.

\begin{figure*}[ht!]
\centering
\scriptsize
\fbox{
\begin{minipage}{0.97\linewidth}
\ttfamily

\noindent\textbf{[INPUT SECTION]} \\
\vspace{0.4em}

\fbox{\textbf{Question:}  
what food contains gluten?}

\vspace{0.6em}
\fbox{\textbf{Reference Passages:}}

Passage 1:  
Foods may also be labeled as gluten-free. 
If a product carries a gluten-free label, ...  

\vspace{0.3em}
Passage 2:  
\hlgreen{Gluten-Containing Grains and Their Derivatives.}  
\hlgreen{Wheat (wheatberries, durum).}  
\hlgreen{Rye. Barley. Triticale.}  
\hlgreen{Malt (malted barley flour, malt extract, malt syrup).}  
\hlgreen{Brewer’s Yeast.}

\vspace{0.3em}
Passage 3:  
Common Foods That Contain Gluten.  
\hlgreen{Pastas: raviolis, dumplings.}  
\hlgreen{Noodles: ramen, udon.}  
\hlgreen{Breads and Pastries: croissants, pita.} 

\vspace{0.8em}
\fbox{\textbf{Sampled Model Response:}}

Based on the provided passages, here are some foods that contain gluten:

1. Wheat: Wheatberries, Durum, \hlred{Whole wheat flour}, \hlred{Bran}, \hlred{Germ}. 

2. Rye: \hlred{Rye berries}, \hlred{Rye flour}.

3. Barley: \hlred{Barley flour}, Malted barley flour, \hlred{Barley sugar}.

...

\hlred{Note: It's important to note that even if a food is listed as containing gluten,  
it may still be safe for consumption for individuals with celiac disease...}

\vspace{0.8em}
\hrule
\vspace{0.8em}
\noindent\textbf{[HALLUCINATION DETECTION OUTPUTS]} \\
\vspace{0.4em}

\fbox{\textbf{Direct LLM Prediction:}}

\{
"hallucinations in response": true, 
 "type": "Evident Introduction of Baseless Information", 
 "explanation": "The 'Note' section at the end is not present in the reference passages."  
\}

\vspace{0.8em}
\hrule
\vspace{0.8em}

\fbox{\textbf{Representative SQL Reasoning:}}

Query the database for all items explicitly marked as containing gluten: \\

\noindent
\texttt{SELECT food\_name, category FROM food WHERE contains\_gluten = TRUE;}

\vspace{0.4em}
\noindent
\fbox{\textbf{SQL Result (excerpt):}}

\begin{tabular}{ll}
\hline
\textbf{food\_name} & \textbf{category} \\
\hline
Wheatberries & Derivative (Wheat) \\
Durum & Derivative (Wheat) \\
Malted barley flour & Derivative (Malt) \\
Raviolis & Common Foods (Pasta) \\
Ramen & Common Foods (Noodles) \\
\hline
\end{tabular}

\vspace{0.4em}
\noindent
\fbox{\textbf{Interpretation:}}  
The \texttt{food} table contains entries derived strictly from the reference text. 
\hlred{Items like 'Whole wheat flour' or 'Barley sugar' are missing from the 
TRUE results}, indicating they are ungrounded hallucinations. The SQL-grounded explanation is much more detailed and incorporates this information.

\vspace{0.8em}
\fbox{\textbf{TeQHallu (ours) Prediction:}}

\{
"hallucinations in response": true,   
 "type": "Evident Introduction of Baseless Information",
"explanation": "\hlorange{The SQL results confirm only high-level grains and select derivatives.} 
\hlorange{The response introduces unsupported specific items: Whole wheat flour, Bran, Germ, 
Rye berries, Rye flour, Barley flour, and Barley sugar.} 
\hlorange{Additionally, no health safety 'Note' exists in the structured database.}"  
\}

\end{minipage}
}
\caption{Full qualitative example from RAGTruth. 
\hlgreen{Green} indicates facts in the reference text; 
\hlred{Red} indicates hallucinations in the sampled response; 
\hlorange{Orange} indicates the detailed SQL-grounded explanation. }
\label{fig:qualitative_example}
\vspace{-10pt}
\end{figure*}

\begin{table}[t]
\centering
\small
\begin{tabular}{l c}
\toprule
\textbf{Task} & \textbf{Parsing Accuracy (\%)} \\
\midrule
Question Answering & 80.00  \\
Data-to-text & 94.00  \\
Summarisation & 82.00  \\
\midrule
\textbf{Overall} & 85.33  \\
\bottomrule
\end{tabular}
\caption{Manual evaluation of SQL parsing accuracy on 50 randomly sampled reference documents per task. Accuracy denotes the percentage of instances where the generated SQL queries fully capture all source text information without data loss.}
\label{tab:sql_parsing_accuracy_manual}
\vspace{-15pt}
\end{table}

\subsection{Qualitative Analysis}

Beyond quantitative results, TeQHallu provides insights through a traceable reasoning chain (\Cref{fig:qualitative_example}).
We employ a multi-colour highlighting scheme where \hlgreen{green} identifies grounded reference facts, \hlred{red} denotes response hallucinations, and \hlorange{orange} highlights the final grounded explanation.
The core strength of our framework lies in a deterministic SQL reasoning step that transforms unstructured reference passages into a symbolic \texttt{food} table for structured verification.
For the ``gluten-containing foods'' query, TeQHallu executes SQL logic to cross-reference claims against the database.
This process reveals that whilst the response correctly identifies primary grains, it hallucinates specific derivatives like ``Bran'' and ``Germ'' which are absent from the source text.
The neurosymbolic checkup then reconciles these SQL results with the initial neural judgement.
In this instance, direct prediction flags the ungrounded health ``Note'' but overlooks the subtle grain-derivative hallucinations.
TeQHallu's final explanation synthesises the symbolic evidence to provide a granular, high-fidelity reasoning trail.
This structured grounding ensures each hallucination label is backed by explicit evidence, providing a transparent justification for human oversight.

\paragraph{SQL Accuracy and Error Propagation Analysis.}
To evaluate grounding fidelity, we manually audit 50 reference documents per task ($150$ total).
As shown in \Cref{tab:sql_parsing_accuracy_manual}, database parsing accuracy is highest in Data-to-text, followed by Summarisation and Question Answering, directly mirroring downstream hallucination detection performance.
Analysing error propagation across the 22 incomplete database cases ($14.67\%$), the pipeline demonstrates strong fallback resilience: 14 instances ($63.6\%$) still yield correct predictions via the reference text safety net in Step~5.
Cascading errors where omitted database records induce incorrect predictions account for only 8 instances ($5.33\%$ total).
Similarly, pure reasoning failures on fully preserved databases occur in only 8 of 128 cases ($6.25\%$).
This confirms that pipeline errors stem primarily from general model reasoning limits rather than misleading SQL representations.

\begin{table}[t!]
\centering
\resizebox{1.\linewidth}{!}{
\small
\begin{tabular}{@{\hspace{0pt}}l@{\hspace{2pt}}|ccc@{\hspace{1pt}}}
\hline
\textbf{Method} & \textbf{Precision} & \textbf{Recall} & \textbf{F1} \\
\hline
Direct Prediction & 48.2 & 95.4 & 61.6 \\
\hline
Symbolic Prediction & 39.9 & 51.3 & 39.7 \\
SELECT Only IDs & 62.4 & 80.1 & 69.7 \\
TeQHallu & 61.7 & 88.9 & \textbf{71.3} \\
\hline
\end{tabular}
}
\caption{Ablation study on the RAGTruth dataset showing overall macro-average precision, recall, and F1.}
\label{tab:ablation}
\vspace{-5pt}
\end{table}

\subsection{Ablation Study}

We conduct an ablation study to isolate the impact of different grounding configurations on TeQHallu's performance.
We compare the baseline neural prediction against a strict intersection over symbolic entity ID sets, which forces deterministic verification but fails to capture the full versatility of hallucinations.
The final symbolic prediction prompt generates SELECT queries for entity IDs to ensure claims in the generated response match the exact identifiers stored within the relational schema.
While relying solely on neural text-to-text comparisons often misses fine-grained entity mismatches, a strict relational comparison proves too rigid.
Instead, evaluating an alternative configuration that feeds these ID sets into the LLM prompt to inform its final decision yields the optimal balance.
This soft LLM-driven alignment leverages the precision of symbolic identifiers without sacrificing the model's contextual reasoning capabilities.

\begin{table}[t!]
\centering
\resizebox{1.\linewidth}{!}{
\scriptsize
\begin{tabular}{@{\hspace{0pt}}l@{\hspace{2pt}}|@{\hspace{2pt}}l@{\hspace{2pt}}|@{\hspace{2pt}}ccc@{\hspace{2pt}}}
\hline
\textbf{Model} & \textbf{Method} & \textbf{Precision} & \textbf{Recall} & \textbf{F1} \\
\hline
\multirow{2}{*}{\textbf{Gemini-2.5-Flash (Spider EM: 51.3)}} 
& Direct Prediction & 48.2 & 95.4 & 61.6 \\
& TeQHallu & 61.7 & 88.9 & 71.3 \\
\hline
\multirow{2}{*}{\textbf{Qwen3 235B/22B (Spider EM: 29.0)}} 
& Direct Prediction & 50.9 & 94.0 & 64.9 \\
& TeQHallu & 68.3 & 67.6 & 67.2 \\
\hline
\multirow{2}{*}{\textbf{GPT-OSS 20B/3.6B (Spider EM: 22.4)}} 
& Direct Prediction & 69.2 & 69.0 & 68.7 \\
& TeQHallu & 64.0 & 74.7 & 68.6 \\
\hline
\multirow{2}{*}{\textbf{LLaMa 3.3 70B (Spider EM: 35.4)}} 
& Direct Prediction & 78.4 & 54.7 & 64.3 \\
& TeQHallu & 76.3 & 57.4 & 65.4 \\
\hline
\end{tabular}
}
\caption{Model hallucination detection results on RAGTruth, reporting macro-average precision, recall, and F1-scores.
Total/active parameters and Spider \cite{yu-etal-2018-spider} multi-turn text-to-SQL exact match (EM) performance are provided in brackets.}
\label{tab:open_model_results}
\vspace{-15pt}
\end{table}

\subsection{Open-Model Results}
\label{experiments:subsec:open}

We evaluate TeQHallu across open-weight models, including Qwen3 \cite{qwen3technicalreport}, GPT-OSS \cite{openai2025gptoss}, and LLaMA~3.3~70B \cite{meta_llama3_3_modelcard_2025}. We also evaluate text-to-SQL exact match performance of these models using the multi-turn setup established by \citet{laban2025llms}, from whom we also source results.
\Cref{tab:open_model_results} summarises macro-averaged performance on RAGTruth across direct prediction and TeQHallu.

Under identical prompting, direct prediction exhibits model-dependent trade-offs: Gemini-2.5-Flash and Qwen3 yield high recall but low precision, whereas LLaMA~3.3~70B prioritises precision.
Integrating SQL grounding via TeQHallu consistently corrects this imbalance across all tested backbones by forcing explicit evidence checks.
Except for GPT-OSS, which possesses the lowest text-to-SQL capability, this low-level symbolic approach yields the most stable performance and highest F1-scores.
This confirms that symbolic grounding and neural judgements are highly complementary, effectively balancing precision and sensitivity.

While backbones with stronger text-to-SQL capabilities yield better overall detection performance, exceptionally high parsing metrics are not required.
Overall, these results show that SQL-grounded detection provides a robust corrective signal regardless of the underlying model.
Whilst smaller architectures like GPT-OSS gain less from the text-to-SQL layer—aligning with \citet{vukovic2025text}—structured evidence consistently enhances reliability across open and closed LLMs.

\subsection{DiaHalu TOD Results}

The response-level hallucination detection results on the task-oriented portion of DiaHalu are summarised in \Cref{tab:diahalu_tod_only}.
Our proposed TeQHallu pipeline consistently achieves the highest F1-scores across all evaluated models, outperforming established baselines and prompt-engineering variants.
Whilst baseline and CoT configurations exhibit high precision, they suffer from low recall, an imbalance our pipeline corrects.
Crucially, TeQHallu maintains this strong performance when constructing the database directly from noisy conversational data, demonstrating robust capabilities on raw dialogues.

For Gemini-2.5-Flash, we also evaluate an alternative configuration (TeQHallu + MWOZ DB) by building a dataset-wide database from MultiWOZ first rather than constructing it incrementally per dialogue.
However, this holistic approach degrades performance because cross-dialogue constraints introduce extra contextual information that makes the verification logic overly strict.

\Cref{tab:fine_hallucination_type_performance} breaks down performance by specific hallucination subtypes, reporting aggregate performance as micro-F1.
TeQHallu drastically improves the detection of Non-factual errors (NF) and substantially boosts Overreliance (Or) detection compared to baseline models.
Incoherence (Ic) detection remains stable across configurations, whilst Irrelevance (Ir) detection exhibits a minor performance decrease.
Despite this minor trade-off, our pipeline achieves the highest overall micro-F1 scores, validating its utility for grounded, conversational verification.

\begin{table}[t]
\centering
\scriptsize
\begin{tabular}{@{\hspace{0pt}}l@{\hspace{2pt}}l@{\hspace{2pt}}c c c@{\hspace{0pt}}}
\toprule
\textbf{Model} & \textbf{Variant} & \textbf{Precision} & \textbf{Recall} & \textbf{F1 Score} \\
\midrule
Random & -- & 31.86 & 48.00 & 38.30 \\
SelfCheckGPT & BertScore & 35.38 & 30.67 & 32.86 \\
 & NLI & 38.84 & 62.67 & 47.96 \\
 & Prompt & 48.00 & 32.00 & 38.40 \\
FOCUS & -- & 34.09 & 60.00 & 43.48 \\
LLaMA-30B & -- & 30.77 & 5.33 & 9.09 \\
Vicuna-33B & -- & 42.86 & 4.00 & 7.32 \\
\cmidrule(l){2-5}
Gemini-1.5 PRO & Base & 60.00 & 36.00 & 45.00 \\
 & w/ CoT & 69.77 & 40.00 & 50.85 \\
 & w/ One-shot & 60.87 & 37.33 & 46.28 \\
 & TeQHallu \textit{(ours)} & 57.33 & 57.33 & \textbf{57.33} \\
\cmidrule(l){2-5}
GPT-4 & Base & 74.19 & 30.67 & 43.40 \\
 & w/ CoT & 73.17 & 40.00 & 51.72 \\
 & w/ One-shot & 71.87 & 30.67 & 42.99 \\
 & TeQHallu \textit{(ours)} & 53.33 & 64.00 & \textbf{58.18} \\
 \cmidrule(l){2-5}
 Gemini-2.5-Flash & Base & 36.87 & 97.33 & 53.48 \\
 & TeQHallu \textit{(ours)} & 58.90 & 57.33 & \textbf{58.11} \\
 & TeQHallu + MWOZ DB & 57.53 & 56.00 & 56.76 \\
\bottomrule
\end{tabular}
\caption{Hallucination detection performance on the DiaHalu TOD portion. Baseline results from \citet{chen-etal-2024-diahalu}. \textbf{Bolding} indicates the best F1-score for each model group using our TeQHallu pipeline.}
\label{tab:diahalu_tod_only}
\vspace{-10pt}
\end{table}

\begin{table}[t]
\centering
\scriptsize
\begin{tabular}{@{\hspace{0pt}}l@{\hspace{2pt}}|@{\hspace{2pt}}cccc@{\hspace{2pt}}|@{\hspace{2pt}}c@{\hspace{0pt}}}
\toprule
 & NF & Ic & Ir & Ov & ALL \\ \midrule
Gemini-1.5 PRO & 27.63 & 42.67 & 23.53 & 30.77 & 31.39 \\
\quad TeQHallu \textit{(ours)} & 45.33 & 50.00 & 11.11 & 40.00 & \textbf{42.38} \\ \midrule
GPT-4 & 33.33 & 44.78 & 0.00 & 0.00 & 33.87 \\
\quad TeQHallu \textit{(ours)} & 40.00 & 45.00 & 0.00 & 40.00 & \textbf{37.13} \\
\midrule
Gemini-2.5-Flash & 27.63 & 42.67 & 23.53 & 30.77 & 31.39 \\
\quad TeQHallu \textit{(ours)} & 48.65 & 47.83 & 10.53 & 40.00 & \textbf{42.95} \\
\bottomrule
\end{tabular}
\caption{Fine-grained hallucination-type recognition F1-scores on the TOD portion. Micro average is reported over all types. \textbf{Bolding} indicates the best overall performance of each model.}
\label{tab:fine_hallucination_type_performance}
\vspace{-15pt}
\end{table}

\subsection{Computational Cost Analysis}

\label{sec:computational_efficiency}

To contextualise the computational footprint of \textsc{TeQHallu}, we evaluate training overhead, inference token volume, parameter scales, and API interaction counts across all comparison paradigms.

\paragraph{Offline Training vs. Zero-Shot Prompting.}
Supervised state-of-the-art baselines such as RAG-HAT \cite{song-etal-2024-rag} and fine-tuned \textsc{Llama-2-13B} \cite{niu-etal-2024-ragtruth} require substantial offline compute budgets, involving full parameter optimisation across thousands of task-specific annotated training pairs. 
Specialised encoders like \textsc{LettuceDetect-large} ($396\text{M}$ parameters) \cite{kovacs-recski-2025-lettucedetect} minimize downstream inference latency, yet remain tethered to specialized pre-annotation and domain-specific retraining. 
Conversely, \textsc{TeQHallu} operates strictly unsupervised with zero training FLOPs ($0\text{ FLOPs}$ offline overhead), performing hallucination detection out-of-the-box on general-purpose instruction-tuned LLMs.

\paragraph{Inference Token Overhead and Prompting Multiplicity.}
While direct prediction baselines require a single forward pass per instance ($1$ API call, consuming $\sim 0.8\text{k}$--$1.8\text{k}$ input tokens), \textsc{TeQHallu} trades context length for symbolic precision. 
Our incremental text-to-SQL construction and verification pipeline requires $5$ sequential prompting steps per document-response pair (inspecting schemas, checking existing records, inserting missing knowledge tuples, querying response claims, and performing final neurosymbolic reconciliation) \cite{vukovic2025text}. 
Cumulatively, carrying prompt state across these $5$ steps incurs an input context overhead of $\sim 4.5\text{k}$--$9.7\text{k}$ tokens per sample, alongside $\sim 600$ generated output tokens. 

\paragraph{Model Scale and Active Parameter Efficiency.}
As demonstrated in \Cref{tab:open_model_results}, \textsc{TeQHallu} does not necessitate dense hundred-billion parameter models to achieve high performance. 
When deployed on Mixture-of-Experts (MoE) backbones such as Gemini 2.5 Flash or Qwen3 ($22\text{B}$ active parameters), the incremental cost per step is mitigated by sparse routing efficiency. 
Even on lightweight open-weight models like GPT-OSS ($3.6\text{B}$ active parameters), \textsc{TeQHallu} achieves stable performance without requiring GPU cluster fine-tuning. 
Thus, \textsc{TeQHallu} effectively shifts the resource burden from expensive offline data curation and parameter optimization to lightweight, inference-time symbolic querying.

\section{Discussion and Future Work}

Evaluations on RAGTruth and DiaHalu demonstrate that grounding hallucination detection within structured SQL databases robustly complements direct neural prediction.
The TeQHallu pipeline achieves competitive F1-scores against zero-shot baselines and fine-tuned models while providing traceable SQL reasoning traces.
Ablation results highlight that using SQL entity IDs directly for prediction creates a tighter symbolic coupling that is too strict to make the final hallucination prediction.
Qualitative analysis confirms that this structured approach identifies subtle hallucinations, such as unsupported grain derivatives, which direct LLM predictions regularly overlook.

A key challenge remains the computational overhead and latency introduced by the multi-stage text-to-SQL generation process.
However, this initial translation overhead yields a substantial downstream advantage by shifting the reasoning burden from fuzzy, autoregressive generation to precise database querying.
This architectural uniformity enables unified cross-domain auditing without requiring task-specific fine-tuning.

Future work will explore integrating knowledge graphs alongside SQL grounding to broaden the scope of verifiable facts.
We also plan to investigate iterative database refinement using detection feedback loops to dynamically improve grounding fidelity over time.
Furthermore, improving performance in predictions based on a a strict neurosymbolic check-up could lead to better interpretability.
Finally, conducting human-in-the-loop evaluations will be essential to validate the utility of TeQHallu's symbolic explanations in high-stakes legal and financial applications.

\section{Conclusion}

We introduce TeQHallu, a neurosymbolic framework that enhances hallucination detection by grounding language model responses in structured SQL databases.
By transforming reference documents into verifiable evidence, our approach matches or exceeds state-of-the-art performance on the RAGTruth and DiaHalu TOD benchmarks.

Beyond quantitative metrics, TeQHallu provides transparent reasoning traces and human-readable explanations via symbolic SQL checkups.
This structured grounding identifies subtle factual inconsistencies that direct neural predictions often overlook, fostering greater trust.

Our findings demonstrate the substantial potential of integrating symbolic reasoning with generative models to ensure factuality.
Future work will focus on expanding the framework to incorporate diverse structured sources and iterative database refinement for high-stakes, knowledge-intensive applications.

\section*{Limitations}

While our pipeline demonstrates strong performance, several limitations exist.

First, effectiveness depends on the quality and completeness of reference documents.
Incomplete or noisy material may cause the SQL-grounded reasoning to miss hallucinations or produce incorrect explanations, although the results on DiaHalu are promising.

Second, constructing and querying the SQL database introduces computational overhead compared to direct prediction.
This may limit applicability in real-time or resource-constrained settings, particularly for large-scale dialogue systems.
Furthermore, there can be an information loss when translating unstructured data to SQL, as shown in our manual analysis.

Third, the pipeline relies on the LLM's ability to generate accurate SQL queries.
Errors in SQL generation can propagate to the final prediction, potentially undermining interpretability.
Furthermore, performance can vary based on exact prompt wording. \citet{vukovic2025text} show that the order in which the documents are presented do not significantly impact the quality of the final ontology.

Finally, while SQL grounding improves explainability, we have not yet conducted large-scale human evaluations.
Such assessments are critical to validate how end-users perceive and utilise these transparency benefits, but are out of scope for this paper.

Addressing these limitations—including exploring hybrid knowledge sources, graded scoring, and human-in-the-loop evaluation—remains a priority for future research.

\section*{Ethical Considerations}

While our framework enhances the reliability of hallucination detection, its deployment involves several ethical considerations.
The pipeline relies on reference documents; if these contain biased, harmful, or non-representative information, the generated SQL database and subsequent hallucination checks will reflect those same biases.
Automated detection may also provide a false sense of security; users might over-rely on the system's ``verified'' status, leading to automation bias in critical domains like healthcare or law.
Furthermore, the computational cost associated with multi-step SQL generation increases the carbon footprint of the inference process, necessitating a balance between detection accuracy and environmental impact.

Finally, transparent reasoning traces could be exploited by malicious actors to reverse-engineer sensitive reference data or to develop adversarial responses that bypass the structured verification logic.
Addressing these risks requires continuous human oversight and robust data governance to ensure that automated fact-checking remains a beneficial tool for digital safety.

\section*{Acknowledgments}

We thank the reviewers for their insightful comments and suggestions, which have substantially improved this paper. We also thank Alexander Koller for his valuable feedback on an earlier draft.

This work was funded by the European Research Council (ERC) under the Horizon 2020 research and innovation programme (Grant No. STG2018 804636) and by the Ministry of Culture and Science of North Rhine-Westphalia through the Lamarr Fellow Network.

Computational resources were provided by the Centre for Information and Media Technology at Heinrich Heine University Düsseldorf and Google Cloud.


\bibliography{custom}

\appendix

\section{Full TeQHallu Prompt}
\label{sec:appendix:prompt}

See \Cref{fig:appendix:full_prompt} for the full TeQHallu Prompt

\begin{figure*}[t!]
    \scriptsize
    \begin{enumerate}

    \item \textbf{Step 1 -- Check existing tables in DB}\\
    You are given a reference.
    Generate only SQL queries (valid for the Python sqlite3 package) to check whether the information from the reference is already present in the database. 
    Here is the current list of tables in the DB: \{db\_result\_input\}.
    First, only generate the \texttt{PRAGMA(table\_info)} queries for the relevant tables. 
    Make sure to highlight SQL using \texttt{```sql ```}.
    Here is the reference: [Reference Text]

    \item \textbf{Step 2 -- Check existing info in DB}\\
    Query results from queries above: \{db\_result\_input\}.
    Generate only SQL queries (valid for the Python sqlite3 package) to check whether the information from the reference is already present in the database. 
    Do this based on the current set of tables and their corresponding columns seen in the \texttt{PRAGMA} query results above. 
    Generate \texttt{SELECT} queries now. 
    Make sure to highlight SQL using \texttt{```sql ```}.

    \item \textbf{Step 3 -- Insert missing info into DB}\\
    Query results from queries above: \{db\_result\_input\}.
    Based on the reference and the results of Step 2, generate only SQL queries to insert the missing information into the database. 
    Do not duplicate existing data. 
    If necessary create tables that are not yet in the DB before adding information. 
    Make sure to highlight SQL using \texttt{```sql ```}.

    \item \textbf{Step 4 -- Query info for response}\\
    You are given a sampled response.
    Generate only SQL queries to retrieve all relevant information from the database that would be necessary to fact-check this response. 
    Make sure to highlight SQL using \texttt{```sql ```}.
    Here is the sampled response: [Response Text]

    \item \textbf{Step 5 -- Detect hallucinations}\\
    Query results from queries above: \{db\_result\_input\}.
    
    Additionally to the context from the steps above you get:
    
    - An initial hallucination prediction, which is a preliminary judgement of whether the response contains hallucinations based on the reference and the response only (this prediction does not use any DB-retrieved information).
    
    Your task is to make a final hallucination determination for the response.
    A hallucination is any information that contradicts or goes beyond what is supported by the reference. Hallucinations fall into one of the following categories:
    
    1. Evident Conflict -- Direct contradiction or opposition to the reference (e.g., factual errors, incorrect numbers, misspelled names).
    
    2. Subtle Conflict -- Divergence from the intended contextual meaning (e.g., substitutions or rephrasings that change implications or severity).
    
    3. Evident Introduction of Baseless Information -- Clear addition of fabricated or unsupported details not found in the reference.
    
    4. Subtle Introduction of Baseless Information -- Inferred details, assumptions, sentiments, or unverifiable insights beyond what the reference states.
    
    Evaluation Process:
    
    1. Consider the initial hallucination prediction as a starting signal.
    
    2. Independently verify the response against: the original reference (primary authority) and the DB-retrieved information (used to confirm or refute factual claims).
    
    3. Assess whether the initial prediction remains accurate after considering the DB information. If the initial prediction is correct, retain it. If the initial prediction is inaccurate or incomplete, update or override it.
    
    4. The final decision must be based on your full evaluation, not solely on the initial prediction.
    Important Guidelines:
    
    - Focus only on information grounded in the reference.
    
    - Do not mark minor wording differences, optional phrases, or harmless stylistic changes as hallucinations.
    
    - Be lenient about paraphrasing and small changes in phrasing.
    
    - Only flag information as a hallucination if it clearly introduces a fact, claim, or detail that is unsupported by or contradicts the reference.
    
    Output Format:
    
    Return your final answer in exactly one of the following JSON formats:
    
    \texttt{\{"hallucinations in response": True, "type": "<category>", "explanation": "<explanation>"\}} or \texttt{\{"hallucinations in response": False\}}.

    \end{enumerate}
    \vspace{-0.5cm}
    \caption{Full prompts for the text-to-SQL hallucination detection pipeline. \texttt{db\_result\_input} contains outputs from preceding SQL executions.}
    \label{fig:appendix:full_prompt}
    \vspace{-10pt}
\end{figure*}


\end{document}